\documentclass[11pt]{article}
\usepackage{acl2016}
\usepackage{times}
\usepackage{url}
\usepackage{latexsym}
\usepackage{amsmath}
\usepackage{multirow}

\usepackage{tikz}
\usetikzlibrary{shapes}
\usepackage{algorithm}
\usepackage{algorithmic}
\usepackage{amssymb}
\usepackage[font=small,labelfont=bf]{caption}

\newcommand{\com}[1]{}
\newcommand{\isection}[2]{\section{#1}\label{ssec:#2}}

\newcommand{\secref}[1]{\textsection{~\ref{ssec:#1}}}

\aclfinalcopy % Uncomment this line for the final submission
\title{Edge-Linear First-Order Dependency Parsing with Undirected Minimum Spanning Tree Inference}

\author{Effi Levi$^1$\\
\And Roi Reichart$^2$\\
          $^1$Institute of Computer Science, The Hebrew Univeristy\\
          $^2$Faculty of Industrial Engineering and Management, Technion, IIT\\
          {\tt \{efle$|$arir\}@cs.huji.ac.il~~roiri@ie.technion.ac.il}\\
\And Ari Rappoport$^1$\\
 }

\date{}

\begin{document}

\maketitle

\begin{abstract}

The run time complexity of state-of-the-art inference algorithms in graph-based dependency parsing  
is super-linear in the number of input words ($n$). Recently, pruning algorithms for these models 
have shown to cut a large portion of the graph edges, with minimal damage to the resulting parse trees. 
Solving the inference problem in run time complexity determined solely 
by the number of edges ($m$) is hence of obvious importance.

We propose such an inference algorithm for first-order models, which encodes the problem as 
a minimum spanning tree (MST) problem in an \textit{undirected graph}.
This allows us to  
utilize state-of-the-art undirected MST algorithms whose run time is $O(m)$ at expectation and with 
a very high probability.
A directed parse tree is then inferred from the undirected MST 
and is subsequently improved with respect to the directed parsing model 
through local greedy updates, both steps running in $O(n)$ time. 
In experiments with 18 languages, a variant of the first-order MSTParser~\cite{McDonald:05b} 
that employs our algorithm performs very similarly to the original parser that 
runs an $O(n^2)$ directed MST inference.

\end{abstract}

\isection{Introduction}{sec:intro}

Dependency parsers are major components of a large number of NLP applications. 
As application models are applied to constantly growing amounts of data, 
efficiency becomes a major consideration.

In graph-based dependency parsing models \cite{Eisner:00,McDonald:05a,McDonald:05b,Carreras:07,Koo:10}, 
given an $n$ word sentence and a model order $k$, the run time of exact inference is  
$O(n^3)$ for $k = 1$ and $O(n^{k+1})$ for $k > 1$ in the projective case~\cite{Eisner:96,McDonald:06}. 
In the non-projective case it is $O(n^2)$ for $k=1$ and NP-hard for $k \geq 2$~\cite{McDonald:07}.~\footnote{We refer to parsing 
approaches that produce only projective dependency trees as projective parsing and 
to approaches that produce all types of dependency trees as non-projective parsing.} 
Consequently, a number of approximated parsers have been introduced, utilizing a variety of techniques: 
the Eisner algorithm \cite{McDonald:06}, 
belief propagation \cite{Smith:08}, dual decomposition \cite{Koo:10,Martins:13} 
and multi-commodity flows \cite{Martins:09,Martins:11}. 
The run time of all these approximations is super-linear in $n$.

Recent pruning algorithms for graph-based dependency 
parsing~\cite{Rush:12,Riedel:12,Zhang:12} have shown to cut a very large portion 
of the graph edges, with minimal damage to the resulting parse trees.
For example, \newcite{Rush:12} demonstrated that a single 
$O(n)$ pass of vine-pruning \cite{Eisner:05} can preserve 
$> 98\%$ of the correct edges, while ruling out  $> 86\%$ of all possible edges.
Such results give strong motivation to solving the inference problem 
in a run time complexity that is determined solely by the number of edges ($m$).~\footnote{Some pruning 
algorithms require initial construction of the full graph, which requires exactly $n(n-1)$ edge weight computations. 
Utilizing other techniques, such as length-dictionary pruning, graph construction and pruning can be 
jointly performed in $O(n)$ steps. 
We therefore do not include initial graph construction and pruning in 
our complexity computations.}

In this paper we propose to formulate the inference problem in first-order (arc-factored) dependency parsing
as a minimum spanning tree (MST) problem in \textit{an undirected graph}. 
Our formulation allows us to employ state-of-the-art algorithms for the MST problem in undirected 
graphs, whose run time depends solely on the number of edges in the graph. 
Importantly,  a parser that employs our undirected inference algorithm 
can generate all possible trees, projective and non-projective.

Particularly, the undirected MST problem (\secref{MST}) has a randomized 
algorithm which is $O(m)$ at expectation and with a very high probability 
(\cite{Karger:95}), as well as an $O(m \cdot \alpha(m,n))$ worst-case deterministic 
algorithm \cite{Pettie:02}, where $\alpha(m,n)$ is a certain natural inverse of Ackermann's 
function \cite{Hazewinkel:01}. As the inverse of Ackermann's function 
grows extremely slowly~\footnote{$\alpha(m,n)$ is less 
than 5 for any practical input sizes $(m,n)$.} the deterministic algorithm is in 
practice linear in $m$ (\secref{state-of-the-art}).
In the rest of the paper we hence refer to the run time of these two algorithms as 
{\it practically linear} in the number of edges $m$. 

Our algorithm has four steps (\secref{parser}). 
First, it encodes the first-order 
dependency parsing inference problem as an undirected MST problem, 
in up to $O(m)$ time.
Then, it computes the MST of the resulting undirected graph. Next, 
it infers a \textit{unique} directed parse tree from the undirected MST.
Finally, the resulting directed tree 
is greedily improved with respect to the directed parsing model. 
Importantly, the last two steps take $O(n)$ time, which makes the total run time
of our algorithm $O(m)$ at expectation and with very high probability.~\footnote{The output 
dependency tree contains exactly $n-1$ edges, therefore $m \geq n-1$, which makes $O(m) + O(n) = O(m)$.}

We integrated our inference algorithm into the first-order parser of~\cite{McDonald:05b}
and compared the resulting parser to the original parser which employs 
the Chu-Liu-Edmonds algorithm (CLE, \cite{Chu:65,Edmonds:67}) for inference. 
CLE is the most efficient exact inference algorithm for graph-based 
first-order non-projective parsers, running at $O(n^2)$ time.\footnote{CLE has faster implementations: 
$O(m+nlogn)$ \cite{Gabow:86} as well as $O(mlogn)$ for sparse graphs \cite{Tarjan:77},
both are super-linear in $n$ for connected graphs.  
We refer here to the classical implementation employed by modern parsers 
(e.g. \cite{McDonald:05b,Martins:13}).}

We experimented (\secref{experiments}) with 17 languages from the 
CoNLL 2006 and 2007 shared tasks on multilingual dependency 
parsing~\cite{buchholz2006conll,nilsson2007conll} and in three English setups. 
Our results reveal that the two algorithms perform very similarly.
While the averaged unlabeled attachment accuracy score (UAS)
of the original parser is 0.97\% higher than ours, 
in 11 of 20 test setups the number of sentences that are better parsed by our parser 
is larger than the number of sentences that are better parsed by the original parser.

Importantly, in this work we present an edge-linear first-order dependency parser which achieves similar accuracy to the existing one, making it an excellent candidate to be used for efficient MST computation in $k$-best trees methods, or to be utilized as an 
inference/initialization subroutine as a part of more complex approximation frameworks such as belief propagation. In addition, our model produces a different solution compared to the existing one (see Table~\ref{head-to-head}), paving the way for using methods such as dual decomposition to combine these two models into a superior one.

Undirected inference has been recently explored 
in the context of transition based parsing \cite{gomez:12,gomez:15}, 
with the motivation of preventing the propagation of erroneous early edge directionality 
decisions to subsequent parsing decisions. Yet, 
to the best of our knowledge this is the first paper to address undirected inference 
for graph based dependency parsing. Our motivation and 
algorithmic challenges are substantially different from those of the earlier transition based work. 

\isection{Undirected MST with the Boruvka Algorithm}{MST}

In this section we define the MST problem in undirected graphs. 
We then discuss the Burovka algorithm~\cite{boruvka:26,nesetril:01} 
which forms the basis for the randomized algorithm of~\cite{Karger:95} we employ in this paper. 
In the next section we will describe the \newcite{Karger:95} algorithm in more details.

\paragraph{Problem Definition.}
For a connected undirected graph $G(V,E)$, where $V$ is the set 
of $n$ vertices and $E$ the set of $m$ weighted edges, the MST problem is defined as finding the 
sub-graph of $G$ which is the tree (a connected acyclic graph) with the 
lowest sum of edge weights. The opposite problem -- finding the maximum 
spanning tree -- can be solved by the same algorithms used for the minimum 
variant by simply negating the graph's edge weights. 

\paragraph{Graph Contraction.}

In order to understand the Boruvka algorithm, let us first define the {\it Graph Contraction} operation. 
For a given undirected graph $G(V,E)$ and a subset $\tilde{E} \subseteq E$, 
this operation creates a new graph, $G_C(V_C,E_C)$. In this new graph, $V_C$ 
consists of a vertex for each connected component in $\tilde{G}(V,\tilde{E})$ 
(these vertices are referred to as \textit{super-vertices}). 
$E_C$, in turn, consists of one edge, $(\hat{u},\hat{v})$, 
for each edge $(u,v) \in E \setminus \tilde{E}$, where $\hat{u},\hat{v} \in V_C$
correspond to $\tilde{G}$'s connected components 
to which $u$ and $v$ respectively belong. 
Note that this definition may result in multiple edges between two vertices in 
$V_C$ (denoted \textit{repetitive edges}) 
as well as in edges from a vertex in $V_C$ to itself (denoted \textit{self edges}).

\begin{algorithm}
\hrule
\vspace{0.1cm}
Contract\_graph
\begin{algorithmic}
	\REQUIRE a graph $G(V,E)$, a subset $\tilde{E}\subseteq E$
	\STATE $C \leftarrow$ connected components of $\tilde{G}(V,\tilde{E})$
	\RETURN $G_C(C,E \setminus \tilde{E})$	
\end{algorithmic}
\hrule
\vspace{0.1cm}
Boruvka-step
\begin{algorithmic}[1]
	\REQUIRE a graph $G(V,E)$
	\FORALL{$(u,v) \in E$}
		\IF{$w(u,v) < w(u.minEdge)$}
			\STATE $u.minEdge \leftarrow (u,v)$
		\ENDIF
		\IF{$w(u,v) < w(v.minEdge)$}
			\STATE $v.minEdge \leftarrow (u,v)$
		\ENDIF
	\ENDFOR
	\FORALL{$v \in V$}
		\STATE $E_m \leftarrow E_m \cup \{v.minEdge\}$
	\ENDFOR
	\STATE $G_B(V_B,E_B) \leftarrow$ {\small Contract\_graph}($G(V,E)$,$E_m$)	
	\STATE Remove from $E_B$ self edges and non-minimal repetitive edges
	\RETURN $G_B(V_B,E_B),E_m$
\end{algorithmic}
\hrule
\caption{The basic step of the Boruvka algorithm for the undirected MST problem.}
\label{alg1}
\end{algorithm}

\begin{figure}
\centerline{
\begin{tabular}{cc}
\includegraphics[scale=0.65]{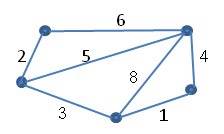} & \includegraphics[scale=0.65]{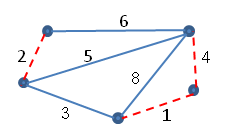} \\
(a) & (b)
\end{tabular}
}
\centerline{
\begin{tabular}{cc}
\includegraphics[scale=0.65]{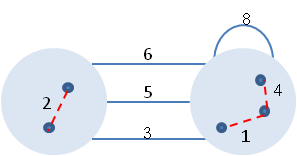} & \includegraphics[scale=0.65]{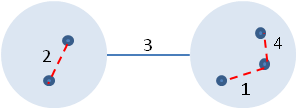} \\
(c) & (d)
\end{tabular}
}
\caption[]{An illustration of a {\it Boruvka step}: 
(a) The original graph; (b) Choosing the minimal edge for each vertex (marked in red); 
(c) The contracted graph; (d) The contracted graph after removing one self edge 
and two non-minimal repetitive edges.} 
\label{Boruvka-example}
\end{figure}

\paragraph{The Boruvka-Step.}

Next, we define the basic step of the Borukva algorithm 
(see example in Figure~\ref{Boruvka-example} and pseudocode in Algorithm \ref{alg1}). 
In each such step, the algorithm creates a subset $E_m \subset E$ by selecting 
the minimally weighted edge for each vertex in the input graph $G(V,E)$
(Figure~\ref{Boruvka-example} (a,b) and Algorithm~\ref{alg1} (lines 1-11)). 
Then, it performs the contraction operation on the graph $G$ and $E_m$ 
to receive a new graph $G_B(V_B,E_B)$ (Figure~\ref{Boruvka-example} (c) and Algorithm~\ref{alg1} (12)). 
Finally, it removes from $E_B$ all self-edges and repetitive 
edges that are not the minimal edges between the vertices $V_B$'s which they connect 
(Figure~\ref{Boruvka-example} (d) and Algorithm~\ref{alg1} (13)). 
The set $E_m$ created in each such step is guaranteed to 
consist only of edges that belong to $G$'s MST and is therefore also returned by the 
Boruvka step.

The Boruvka algorithm runs successive Boruvka-steps until it is left with 
a single super-vertex. The MST of the original graph $G$ is given by the 
unification of the $E_m$ sets returned in each step. The resulting computational complexity 
is $O(m\log{n})$~\cite{nesetril:01}.
We now turn to describe how the undirected MST problem can
be solved in a time practically linear in the number of graph edges.

\isection{Undirected MST in Edge Linear Time}{state-of-the-art}

There are two algorithms that solve the undirected MST problem 
in time practically linear in the number of edges in the input graph. 
These algorithms are based on substantially different approaches: 
one is deterministic and the other is randomized 
\footnote {Both these algorithms deal with a slightly more general case where the 
graph is not necessarily connected, in which case the 
\textit{minimum spanning forest (MSF)} is computed. 
In our case, where the graph is connected, the MSF reduces 
to an MST. }. 

The complexity of the first, deterministic, 
algorithm~\cite{Chazelle:00a,Pettie:02} is $O(m \cdot \alpha(m,n))$, where $\alpha(m,n)$ 
is a natural inverse of Ackermann's function, whose value for any 
practical values of $n$ and $m$ is lower than 5. 
As this algorithm employs very complex data-structures, we 
do not implement it in this paper. 

The second, randomized, algorithm~\cite{Karger:95} has an expected run time of 
$O(m+n)$ (which for connected graphs is $O(m)$), and this run time is achieved with 
a high probability of $1-exp(-\Omega(m))$.~\footnote{This complexity analysis  
is beyond the scope of this paper.}  
In this paper we employ only 
this algorithm for first-order graph-based parsing inference, and hence  
describe it in details in this section. 

\paragraph {Definitions and Properties.}
We first quote two properties of undirected graphs~\cite{Tarjan:83}: (1) The {\bf cycle property}: The 
heaviest edge in a 
cycle in a graph does not appear in the MSF; and 
(2) The {\bf cut property}: For any proper nonempty subset $V'$ of 
the graph vertices, the lightest edge with exactly one endpoint in $V'$ 
is included in the MSF.

We continue with a number of definitions and observations. 
Given an undirected graph $G(V,E)$ with weighted edges, and a forest $F$ in that 
graph, $F(u,v)$ is the path in that forest between $u$ and $v$ (if such a path exists), 
and $s_F(u,v)$ is the maximum weight of an edge in $F(u,v)$ 
(if the path does not exist then $s_F(u,v) = \infty$). 
An edge $(u,v) \in E$ is called {\it F-heavy} if $s(u,v) > s_F(u,v)$, 
otherwise it is called {\it F-light}. 
An alternative equivalent definition is that an edge is {\it F-heavy} 
if adding it to $F$ creates a cycle in which it is the heaviest edge. 
An important observation (derived from the cycle property) 
is that for any forest $F$, no {\it F-heavy} edge can possibly be a part of an MSF for $G$. 
It has been shown that given a forest $F$, 
all the {\it F-heavy} edges in $G$ can be found in $O(m)$ time~\cite{Dixon:92,King:95}.

\begin{algorithm}[t!]
\hrule
\vspace{0.1cm}
Randomized\_MSF
\begin{algorithmic}[1]
	\REQUIRE a graph $G(V,E)$
	\IF{$E$ is empty}
		\RETURN $\emptyset$
	\ENDIF
	\STATE $G_C(V_C,E_C),E_m \leftarrow$ Boruvka-step2($G$)
	\FORALL{$(u,v) \in E_C$}
		\IF{coin-flip == head}
			\STATE $E_s \leftarrow E_s \cup \{(u,v)\}$
			\STATE $V_s \leftarrow V_s \cup \{u,v\}$
		\ENDIF
	\ENDFOR
	\STATE $F \leftarrow$ Randomized\_MSF($G_s(V_s,E_s)$)
	\STATE remove all {\it F-heavy} edges from $G_C(V_C,E_C)$
	\STATE $F_C \leftarrow$ Randomized\_MSF($G_C(V_C,E_C)$)
	\RETURN $F_C \cup E_m$
\end{algorithmic}
\hrule
\caption {Pseudocode for the Randomized MSF algorithm of\protect\cite{Karger:95}.}
\label{alg2}
\end{algorithm}

\paragraph {Algorithm.}

The randomized algorithm can be outlined as follows 
(see pseudocode in algorithm~\ref{alg2}): 
first, two successive Boruvka-steps are applied to the graph 
(line 4, \textit{Boruvka-step2} stands for two successive Boruvka-steps), 
reducing the number of vertices by (at least) 
a factor of 4 to receive a contracted graph $G_C$ and an edge set
$E_m$ (\secref{MST}). 
Then, a subgraph $G_s$ is randomly constructed, such that each edge in
$G_C$, along with the vertices which it connects, is included in $G_s$ 
with probability $\frac{1}{2}$ (lines 5-10).
Next, the algorithm is recursively applied to $G_s$ to obtain 
its minimum spanning forest $F$ (line 11). 
Then, all {\it F-heavy} edges are removed from $G_C$ (line 12), 
and the algorithm is recursively applied to the resulting graph to obtain 
a spanning forest $F_C$ (line 13). The union of that forest 
with the edges $E_m$ forms the requested spanning forest (line 14). 
 
\paragraph{Correctness.}

The correctness of the algorithm is proved by induction. 
By the cut property, every edge returned by the Boruvka step (line 4), 
is part of the MSF. Therefore, the rest of the edges in the original 
graph's MSF form an MSF for the contracted graph. 
The removed {\it F-heavy} edges are, by the cycle property, not part of 
the MSF (line 12). By the induction assumption, the MSF of the remaining graph 
is then given by the second recursive call (line 13).

\isection{Undirected MST Inference for Dependency Parsing}{parser}
 
There are several challenges in the construction of an \textit{undirected MST parser}:  
an MST parser that employs an undirected MST algorithm for inference.\footnote{Henceforth, 
we refer to an MST parser that employs a directed MST algorithm for inference 
as \textit{directed MST parser}.}
These challenges stem from the mismatch between the undirected nature of the inference algorithm and the 
directed nature of the resulting parse tree.

The first problem is that of \textit{undirected encoding}. 
Unlike directed MST parsers that explicitly encode the directed nature of dependency parsing into a 
directed input graph to which an MST algorithm is applied \cite{McDonald:05b}, 
an undirected MST parser needs to encode directionality information into an undirected graph.
In this section we consider two solutions to this problem. 
 
The second problem is that of \textit{scheme conversion}. 
The output of an undirected MST algorithm 
is an undirected tree while the dependency parsing problem requires finding 
a directed parse tree. In this section 
we show that for rooted undirected spanning trees there is only one way to 
define the edge directions under the 
constraint that the root vertex has no incoming edges and that each non-root 
vertex has exactly one incoming edge in the resulting  
directed spanning tree. As dependency parse trees obey 
the first constraint and the second constraint is a definitive property 
of directed trees, 
the output of an undirected MST parser can be transformed 
into a directed tree using a simple $O(n)$ time procedure.

Unfortunately, as we will see in \secref{experiments}, even with our best 
undirected encoding method, an undirected MST 
parser does not produce directed trees of the same quality as its directed counterpart. 
At the last part of this section we therefore present a simple, $O(n)$ time, 
\textit {local enhancement procedure}, that improves the score of the 
directed tree generated from the output of the undirected MST parser with respect to the 
edge scores of a standard directed MST parser. 
That is, our procedure improves the output of the undirected MST parser 
with respect to a directed model without having to compute the MST of the latter, 
which would take $O(n^2)$ time.

We conclude this section with a final remark stating that the output class of our inference algorithm 
is non-projective. That is, it can generate all possible parse trees, projective and non-projective.

\paragraph{Undirected Encoding}

Our challenge here is to design an encoding scheme that encodes 
directionality information into the graph of the undirected MST problem. 
One approach would be to compute directed edge weights according to a feature 
representation scheme for directed edges 
(e.g. one of the schemes employed by existing directed MST parsers) and then 
transform these directed weights into undirected ones. 

Specifically, given two vertices $u$ and $v$ with directed edges $(u,v)$ and $(v,u)$, 
weighted with $s_d(u,v)$ and $s_d(v,u)$ respectively, the goal is to compute the weight 
$s_u(\hat{u,v})$ of the undirected edge $(\hat{u,v})$ connecting them in the undirected graph. We 
do this using a pre-determined function $f:\mathbb{R} \times \mathbb{R} \rightarrow \mathbb{R}$, such that 
$f(s_d(u,v),s_d(v,u)) = s_u(\hat{u,v})$. 
$f$ can take several forms including mean, product and so on.
In our experiments the mean proved to be the best choice.

Training with the above approach is implemented as follows. $w$, the parameter 
vector of the parser, consists of the weights of \textit{directed features}. 
At each training iteration, $w$ is used for the computation 
of $s_d(u,v) = w \cdot \phi(u,v)$ and $s_d(v,u) = w \cdot \phi(v,u)$ 
(where $\phi(u,v)$ and $\phi(v,u)$ are the feature representations of these directed edges). 
Then, $f$ is applied to compute the undirected edge score $s_u(\hat{u,v})$. 
Next, the undirected MST algorithm is run on the resulting weighted undirected 
graph, and its output MST is transformed into a directed tree (see below). 
Finally, this directed tree is used for the update of $w$ with 
respect to the gold standard (directed) tree. 

At test time, the vector $w$ which resulted from the training process 
is used for $s_d$ computations. Undirected graph construction, undirected MST computation 
and the undirected to directed tree conversion process are conducted exactly as in training.~\footnote
{In evaluation setup experiments we also considered a variant of this model where the training 
process utilized directed MST inference. As this variant performed 
poorly, we exclude it from our discussion in the rest of the paper.}

Unfortunately, preliminary experiments in our development setup revealed 
that this approach yields parse trees of much lower quality 
compared to the trees generated by the directed MST parser that employed the original directed feature set. 
In \secref{experiments} we discuss these results in details.

An alternative approach is to employ an undirected feature set. To implement this 
approach, we employed the feature set of the MST parser 
(\cite {McDonald:05a}, Table 1) with one difference: some of the features are directional, 
distinguishing between the properties of the source (parent) and the target (child) vertices. 
We stripped those features from that information, which resulted in an undirected version of the feature set.

Under this feature representation, training with undirected inference is simple. 
$w$, the parameter vector of the parser, now consists of the weights of 
\textit{undirected features}. Once the undirected MST is computed by an undirected  
MST algorithm, $w$ can be updated with respect to an undirected variant of the gold parse trees. 
At test time, the algorithm constructs an undirected graph using the vector $w$ resulted from 
the training process. This graph's undirected MST is computed and then 
transformed into a directed tree.  

Interestingly, although this approach does not explicitly encode edge directionality information 
into the undirected model, it performed very well in our experiments 
(\secref{experiments}), especially when combined with 
the local enhancement procedure described below. 

\begin{figure}
\centerline{
\begin{tabular}{cccc}
\includegraphics[scale=0.8]{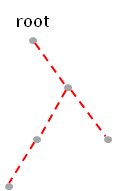} & \includegraphics[scale=0.8]{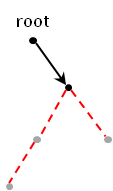} & \includegraphics[scale=0.8]{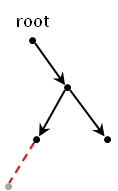} & \includegraphics[scale=0.8]{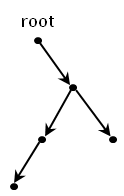} \\
(a) & (b) & (c) & (d) 
\end{tabular}
}
\caption[]{An illustration of directing an undirected tree, given a constrained root vertex: 
(a) The initial undirected tree; 
(b) Directing the root's outgoing edge; 
(c) Directing the root's child's outgoing edges; 
(d) Directing the last remaining edge, resulting in a directed tree.}
\label{directing-fig}
\end{figure}

\begin{figure}
\centerline{
\begin{tabular}{ccc}
\includegraphics[scale=0.6]{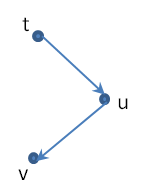} & \includegraphics[scale=0.6]{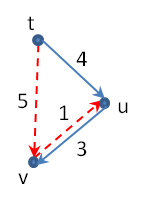} & \includegraphics[scale=0.6]{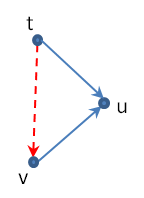} \\
(a) & (b) & (c)
\end{tabular}
}
\centerline{
\begin{tabular}{cc}
\includegraphics[scale=0.6]{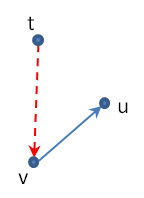} & \includegraphics[scale=0.6]{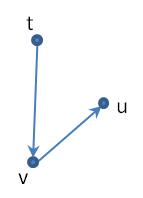}\\
(d) & (e)
\end{tabular}
}
\caption{An illustration of the local enhancement procedure 
for an edge $(u,v)$ in the du-tree. 
Solid lines indicate edges in the du-tree, 
while dashed lines indicate edges not in the du-tree. 
(a) Example subtree; 
(b) Evaluate $gain = s_d(t,u) + s_d(u,v) - (s_d(t,v) + s_d(v,u)) = 4+3-(5+1) = 1$; 
(c) In case a modification is made, first replace 
$(u,v)$ with $(v,u)$; and then 
(d) Remove the edge $(t,u)$; and, finally, 
(e) Add the edge $(t,v)$.} 
\label{trick-fig}
\end{figure}

\paragraph{Scheme Conversion}

Once the undirected MST is found, we need to direct its edges in order for the end result to be a 
directed dependency parse tree. 
Following a standard practice in graph-based dependency parsing (e.g.~\cite{McDonald:05b}),
before inference is performed we add a dummy root vertex to the initial input graph 
with edges connecting it to all of the other vertices in the graph.
Consequently, the final undirected tree will have a designated root vertex. 
In the resulting directed tree, this vertex is constrained to have only outgoing edges. As observed by \newcite{gomez:12}, 
this effectively forces the direction for the rest of the edges in the tree. 

Given a root vertex that follows the above constraint, and together with the definitive property of directed trees 
stating that each non-root vertex in the graph has exactly one incoming edge, 
we can direct the edges of the undirected tree using a simple BFS-like algorithm (Figure~\ref{directing-fig}). 
Starting with the root vertex, we mark its undirected edges as outgoing, 
mark the vertex itself as \textit{done} and its descendants as \textit{open}. 
We then recursively repeat the same procedure for each open vertex until there 
are no such vertices left in the tree, at which point we have a directed tree. 
Note that given the constraints on the root vertex, there is no other way to 
direct the undirected tree edges. This procedure 
runs in $O(n)$ time, as it requires a constant number of operations for each 
of the $n-1$ edges of the undirected spanning tree.

In the rest of the paper we refer to the \emph{directed} tree generated by 
the undirected and directed MST parsers as \textit{du-tree} and 
\textit{dd-tree} respectively.

\paragraph{Local Enhancement Procedure}

As noted above, experiments in our development setup 
(\secref{experiments}) revealed that the directed parser  
performs somewhat better than the undirected one. 
This motivated us to develop a local enhancement procedure that 
improves the tree produced by the undirected model with respect 
to the directed model without compromising our $O(m)$ run time.
Our enhancement procedure is motivated by development 
experiments, revealing the much smaller gap between the quality of the du-tree and dd-tree 
of the same sentence under undirected evaluation compared to directed evaluation 
(\secref{experiments} demonstrates this for test results).

For a du-tree that contains the vertex $u$ and the edges $(t,u)$ 
and $(u,v)$, we therefore consider the replacement of $(u,v)$ with $(v,u)$. 
Note that after this change our graph would no longer be a directed tree, 
since it would cause $u$ to have two parents, $v$ and $t$, and $v$ to have 
no parent. This, however, can be rectified by replacing the edge $(t,u)$  
with the edge $(t,v)$. 

It is easy to infer whether this change results in a better (lower weight) spanning 
tree under the directed model by computing the equation: 
$gain = s_d(t,u) + s_d(u,v) - (s_d(t,v) + s_d(v,u))$, where $s_d(x,y)$ 
is the score of the edge $(x,y)$ according to the directed model. 
This is illustrated in Figure~\ref{trick-fig}.

Given the du-tree, we traverse its edges and compute the above gain for each. 
We then choose the edge with the maximal positive gain, 
as this forms the maximal possible decrease in the directed model score using modifications 
of the type we consider, and perform the corresponding modification. In our experiments we performed this 
procedure five times per inference problem.\footnote{This hyperparameter was estimated once on our 
English development setup, and used for all 20 multilingual test setups.}
This procedure performs a constant number of operations 
for each of the $n-1$ edges of the du-tree, resulting in $O(n)$ run time.

\paragraph {Output Class.}

Our undirected MST parser is non-projective. 
This stems from the fact that the undirected MST algorithms we discuss in \secref{state-of-the-art} 
do not enforce any structural constraint, and particularly the non-crossing constraint, 
on the resulting undirected MST. 
As the scheme conversion (edge directing) and the local enhancement procedures described in this section do 
not enforce any such constraint as well, the resulting tree can take any possible structure.

\isection{Experiments and Results}{experiments}

\begin{table*}
\scriptsize
\centering
\begin{tabular}{ c|c|c|c|c|c|c|c|c|c|c }
   & Swedish & Danish & Bulgarian & Slovene & Chinese & Hungarian & Turkish & German & Czech & Dutch \\
   \hline
		D-MST	        & 87.7/88.9	& 88.5/89.5	& 90.4/90.9	& 80.4/83.4	& 86.1/87.7	& 82.9/84.3	& 75.2/75.3	& 89.6/90.2	& 81.7/84.0	& 81.3/83.0 \\
		\hline
		U-MST-uf-lep	& 86.9/88.4	& 87.7/88.9	& 89.7/90.6	& 79.4/82.8	& 84.8/86.7	& 81.8/83.3	& 74.9/75.3	& 88.7/89.5	& 79.6/82.5	& 78.7/80.7 \\
		\hline
		U-MST-uf	& 84.3/87.8	& 85.1/89.0	& 87.0/90.2	& 76.1/82.4	& 81.1/86.4	& 79.9/82.9	& 73.1/75.0	& 86.9/89.0	& 76.1/81.9	& 73.4/80.5 \\
		\hline
		U-MST-df	& 72.0/79.2	& 74.3/82.9	& 69.5/81.4	& 66.8/75.8	& 65.9/76.5	& 68.2/72.1	& 57.4/62.6	& 77.7/82.5	& 57.3/70.9	& 59.0/71.3 \\
   \hline
	 \hline
	                          & Japanese & Spanish & Catalan & Greek & Basque & Portuguese & Italian & PTB & QBank & GENIA \\
   \hline
                D-MST	        & 92.5/92.6	& 83.8/86.0	& 91.8/92.2	& 82.7/84.9	& 72.1/75.8	& 89.2/89.9   & 83.4/85.4	& 92.1/92.8	& 95.8/96.3	& 88.9/90.0\\
		\hline
		U-MST-uf-lep	& 92.1/92.2	& 83.5/85.9	& 91.3/91.9	& 81.8/84.4	& 71.6/75.8	& 88.3/89.3   & 82.4/84.7	& 90.6/91.7	& 95.6/96.2	& 87.2/88.9 \\
		\hline
		U-MST-uf	& 91.4/92.4	& 80.4/85.4	& 89.7/91.7	& 78.7/84	& 68.8/75.4	& 85.8/89.3   & 79.4/84.4	& 88.5/91.8	& 94.8/96.0	& 85.0/89.0 \\
		\hline
		U-MST-df	& 74.4/85.2	& 73.1/81.3	& 73.1/83.5	& 71.3/78.7	& 62.8/71.4	& 67.9/79.7   & 65.2/77.2	& 77.2/85.4	& 89.1/92.9	& 72.4/81.6 \\
         \hline	
         \hline
\end{tabular}
\caption{\label{results_directed} Directed/undirected UAS for the various parsing models of this paper.}
\end{table*}

\paragraph{Experimental setup}

\begin{table*}[t!]
\scriptsize
\centering
\begin{tabular}{ c|c|c|c|c|c|c|c|c|c|c }
        & Swedish & Danish & Bulgarian & Slovene & Chinese & Hungarian & Turkish & German & Czech & Dutch \\
\hline
D-MST & 20.6 & 20.8 & 15.1 & 25.4 & 15.5 & 26.4 & 22.3 & 21.3 & 29.7 & 27.7 \\
\hline
U-MST-uf-lep & 18.0 & 24.5 & 22.1 & 29.6 & 16.7 & 27.2 & 19.3 & 17.9 & 26.2 & 24.4 \\
\hline
Oracle & 88.9   & 89.7   & 91.6   & 81.9   & 87.8   & 83.9   & 77.1   & 90.6 & 82.8   & 82.8 \\
       & (+1.2) & (+1.4) & (+1.2) & (+1.5) & (+1.7) & (+1) & (+1.9) & (+1) & (+1.1) & (+1.5) \\
\hline
\hline
        & Japanese & Spanish & Catalan & Greek & Basque & Portuguese & Italian & PTB & QBank & GENIA \\
\hline
D-MST & 5.7 & 26.7 & 23.4 & 28.9 & 23.4 & 22.6 & 22.5 & 27.8 & 5.3 & 33.7 \\
\hline
U-MST-uf-lep & 4.0 & 30.1 & 26.3 & 30.5 & 30.8 & 21.9 & 24.9 & 20.9 & 6.0 & 23.8 \\
\hline
Oracle & 93.1   & 84.8 & 92.6   & 83.9   & 74.1  & 89.9   & 84.4   & 92.8   & 96.4   & 89.7   \\
       & (+0.6) & (+1) & (+0.8) & (+1.2) &(+2) & (+0.7) & (+1) & (+0.7) & (+0.8) & (+0.8)  \\
\hline
\end{tabular}
\caption{\label{head-to-head} Top two lines (per language): percentage of sentences for which each of the 
models performs better than the other according to the directed UAS. 
Bottom line (Oracle): Directed UAS of an oracle model that selects the parse tree of 
the best performing model for each sentence. Improvement over the directed UAS score of D-MST
is given in parenthesis.}
\end{table*}

We evaluate four models:
(a) The original directed parser (D-MST, \cite{McDonald:05b}); 
(b) Our undirected MST parser with undirected features and with 
the local enhancement procedure (U-MST-uf-lep);\footnote{The directed edge weights 
for the local enhancement procedure ($s_d$ in \secref{parser}) 
were computed using the trained D-MST parser.} 
(c) Our undirected MST parser with undirected features but without 
the local enhancement procedure (U-MST-uf); and 
(d) Our undirected MST parser with directed features (U-MST-df).
All models are implemented within the MSTParser 
code\footnote{\url{http://www.seas.upenn.edu/~strctlrn/MSTParser/MSTParser.html}}.

The MSTParser does not prune its input graphs. 
To demonstrate the value of undirected parsing for sparse input graphs, 
we implemented the length-dictionary pruning strategy 
which eliminates all edges longer than the maximum length observed for each \textit{directed} 
head-modifier POS pair in the training data. An undirected 
edge $\hat{(u,v)}$ is pruned $iff$ both directed edges $(u,v)$ and $(v,u)$ 
are to be pruned according to the pruning method. 
To estimate the accuracy/graph-size tradeoff provided by undirected parsing (models (b)-(d)), 
we apply the pruning strategy only to these models leaving the the D-MST model (model (a)) untouched. 
This way D-MST runs on a complete directed graph with $n^2$ edges.

Our models were developed in a monolingual setup: training on sections 2-21 of  WSJ 
PTB~\cite{marcus:93} and testing on section 22. 
The development phase was devoted to the various decisions detailed throughout this paper 
and to the tuning of the single hyperparameter: the number of times the local enhancement procedure 
is executed.

We tested the models in 3 English and 17 multilingual setups. The English setups are:  
(a) \textit{PTB}: training on sections 2-21 of the 
WSJ PTB and testing on its section 23; 
(b) \textit{GENIA}: training with a random sample of 90\% of the 4661 
GENIA corpus~\cite{ohta2002genia} sentences 
and testing on the other 10\%; 
and (c) \textit{QBank}: a setup identical to (b) for the 3987 
QuestionBank~\cite{judge2006questionbank} sentences.
Multilingual parsing was performed with the multilingual datasets of the 
CoNLL 2006~\cite{buchholz2006conll} and 2007~\cite{nilsson2007conll} 
shared tasks on multilingual dependency parsing, following their standard 
train/test split. Following previous work, punctuation was excluded from the 
evaluation.

Length-dictionary pruning reduces the number of undirected edges by 27.02\% on average across 
our 20 setups (std = 11.02\%, median = 23.85\%), leaving an average of 73.98\% of the edges in the undirected graph.
In 17 of 20 setups the reduction is above 20\%. Note that the number of edges in 
a complete directed graph is twice the number in its undirected counterpart. 
Therefore, on average, the number of input edges in the pruned undirected models 
amounts to $\frac{73.98\%} {2} = 36.49\%$ of the number of edges in the complete directed graphs. 
In fact, every edge-related operation (such as feature extraction) in the undirected model is actually
performed on half of the number of edges compared to the directed model, saving run-time not only
in the MST-inference stage but in every stage involving these operations. In addition, some pruning
methods, such as length-dictionary pruning (used in this work) perform feature extraction only for existing
(un-pruned) edges, meaning that any reduction in the number of edges also reduces feature extraction operations.

For each model we report the standard directed unlabeled attachment accuracy score (D-UAS). 
In addition, since this paper explores the value of undirected inference for 
a problem that is directed in nature, we also report the 
undirected unlabeled attachment accuracy score (U-UAS), hoping that these 
results will shed light on the differences between the trees generated by the different models.

\paragraph{Results}

Table~\ref{results_directed} presents our main results.
While the directed MST parser (D-MST) is the best performing model 
across almost all test sets and evaluation measures, it outperforms our best model, 
U-MST-uf-lep, by a very small margin.

Particularly, for D-UAS, D-MST outperforms U-MST-uf-lep by up to 
1\% in 14 out of 20 setups (in 6 setups the difference is up to 0.5\%). 
In 5 other setups the difference between the models is between 1\% and 2\%, and only 
in one setup it is above 2\% (2.6\%). Similarly, for  
U-UAS, in 2 setups the models achieve the same performance, in 
15 setups the difference is less than 1\% and in the other setups the differences is 1.1\% - 1.5\%. 
The average differences are 0.97\% and 0.67\% for D-UAS and U-UAS respectively.

The table further demonstrates the value of the local enhancement procedure. 
Indeed, U-MST-uf-lep outperforms U-MST in all 20 setups in D-UAS evaluation 
and in 15 out of 20 setups in U-UAS evaluation 
(in one setup there is a tie). However, the improvement this procedure provides 
is much more noticeable for D-UAS, with an averaged improvement 
of 2.35\% across setups, compared to an averaged U-UAS improvement 
of only 0.26\% across setups. 
While half of the changes performed by the local enhancement procedure are in edge directions, 
its marginal U-UAS improvement indicates that almost all of its power 
comes from edge direction changes. This calls for an improved enhancement procedure.

Finally, moving to directed features (the U-MST-df model), both D-UAS 
and U-UAS substantially degrade, with more noticeable 
degradation in the former. We hypothesize that this stems from the idiosyncrasy 
between the directed parameter update and the undirected inference in this model.

Table~\ref{head-to-head} reveals the complementary nature of our 
U-MST-uf-lep model and the classical D-MST:  
each of the models outperforms the other on an average of 22.2\% of the sentences across test setups.
An oracle model that selects the parse tree of the best model for each sentence would improve 
D-UAS by an average of 1.2\% over D-MST across the test setups. 

The results demonstrate the power of first-order graph-based dependency parsing with 
undirected inference. Although using a substantially different inference algorithm, 
our U-MST-uf-lep model performs very similarly to the standard MST parser which 
employs directed MST inference.

\isection{Discussion}{discussion}

We present a first-order graph-based dependency parsing model which runs in edge linear time 
at expectation and with very high probability.
In extensive multilingual experiments our model performs very similarly to a standard directed 
first-order parser. Moreover, our results demonstrate the complementary nature of the models, with our model 
outperforming its directed counterpart on an average of 22.2\% of the test sentences.

Beyond its practical implications, our work provides a novel intellectual contribution in demonstrating 
the power of undirected graph based methods in solving an NLP problem that is directed in nature. 
We believe this contribution has the potential to affect future research on additional NLP problems.

The potential embodied in this work extends to a number of promising research directions:
\begin{itemize}
\item Our algorithm may be used for efficient MST computation in $k$-best trees methods which are instrumental in
margin-based training algorithms. For example, \newcite{McDonald:05b} observed that $k$ calls to the CLU algorithm might
prove to be too inefficient; our more efficient algorithm may provide the remedy.
\item It may also be utilized as an inference/initialization subroutine as a part of more complex approximation frameworks 
such as belief propagation (e.g. \newcite{Smith:08}, \newcite{Gormley:15}).
\item Finally, the complementary nature of the directed and undirected parsers motivates the development of methods for 
their combination, such as dual decomposition (e.g. \newcite{Rush:10}, \newcite{Koo:10a}). Particularly, we have shown that our undirected inference algorithm converges to a different solution than the standard directed solution while still maintaining high quality (Table~\ref{head-to-head}). Such techniques can exploit this diversity to produce a higher quality unified solution.
\end{itemize}

We intend to investigate all of these directions in future work. In addition, we are currently exploring potential extensions of the techniques presented in this paper to higher order, projective and non-projective, dependency parsing.

\section*{Acknowledgments}
The second author was partly supported by a GIF Young Scientists' Program   grant No. I-2388-407.6/2015 - Syntactic Parsing in Context.

\bibliographystyle{acl2016}
\bibliography{acl2016}

\end{document}